\documentclass{article}

\usepackage[preprint]{neurips_2024}

\usepackage[utf8]{inputenc} 
\usepackage[T1]{fontenc}    
\usepackage{url}            
\usepackage{booktabs}       
\usepackage{amsfonts}       
\usepackage{nicefrac}       
\usepackage{microtype}      
\usepackage{xcolor}         
\usepackage{threeparttable}
\usepackage{ulem}
\usepackage{amsmath}
\usepackage{algorithm}
\usepackage{algorithm}  
\usepackage{algorithmicx}  
\usepackage{algpseudocode}
\usepackage{mdframed}
\usepackage{graphicx} 
\usepackage{amssymb}
\usepackage{multirow}
\usepackage{tabularx}
\usepackage{booktabs}
\usepackage{geometry}
\usepackage{longtable}
\usepackage{times,a4wide}
\usepackage{graphicx}
\usepackage{amsmath}
\usepackage{listings}
\usepackage{comment}
\usepackage{hyperref} 
\usepackage{cleveref}

\usepackage[textsize=tiny]{todonotes}

\usepackage{listings}
\usepackage{float}

\title{SEKI: Self-Evolution and Knowledge Inspiration based Neural Architecture Search via Large Language Models}
\newcommand*\samethanks[1][\value{footnote}]{\footnotemark[#1]}

\author{
    Zicheng Cai\thanks{Equal contribution. tangyaohua28@gmail.com}  \; \; 
    Yaohua Tang\samethanks{}  \; \; 
    Yutao Lai \; \; 
    Hua Wang 
    \\
    \textbf{
    Zhi Chen \thanks{Corresponding author. zhic@mthreads.com} \; 
    Hao Chen \; } 
    \\ \\ 
    \textbf{Moore Threads AI} 
    \\
    \textbf{GuangDong University of Technology} 
}
\begin{document}
\maketitle
\begin{abstract}
We introduce SEKI, a novel large language model (LLM)-based neural architecture search (NAS) method. Inspired by the chain-of-thought (CoT) paradigm in modern LLMs, SEKI operates in two key stages: self-evolution and knowledge distillation. In the self-evolution stage, LLMs initially lack sufficient reference examples, so we implement an iterative refinement mechanism that enhances architectures based on performance feedback. Over time, this process accumulates a repository of high-performance architectures. In the knowledge distillation stage, LLMs analyze common patterns among these architectures to generate new, optimized designs. Combining these two stages, SEKI greatly leverages the capacity of LLMs on NAS and without requiring any domain-specific data. Experimental results show that SEKI achieves state-of-the-art (SOTA) performance across various datasets and search spaces while requiring only 0.05 GPU-days, outperforming existing methods in both efficiency and accuracy. Furthermore, SEKI demonstrates strong generalization capabilities, achieving SOTA-competitive results across multiple tasks.

\end{abstract}

\section{Introduction}
Designing high-performance deep neural network architectures requires significant human efforts and extensive experimentation. To accelerate the development of neural networks, Neural Architecture Search (NAS) has been introduced as an automated approach to efficiently and cost-effectively identify optimal network designs.
In the early stages, methods like NASNet-A~\cite{2017Learning} and AmoebaNet-B~
\cite{2018Regularized} brought the concept of automated architecture search to life, despite their high computational costs of 3150 and 1800 GPU-days, respectively. Later, gradient-based methods~\cite{liu2018darts,xu2019pc,xiao2022shapley} set a new trend by leveraging weight sharing and continuous relaxation techniques, reducing the search cost to as low as 0.4 GPU-days. However, the applicability of these methods was hindered by issues like unfair operation selection, which often resulted in performance collapse. At the same time, evolution-based methods such as EPCNAS-C~\cite{9930866} and EAEPSO~\cite{10045029} achieved significant improvements in efficiency but continued to face challenges in delivering high performance. Additionally, training-free approaches like PINAT~\cite{lu2023pinat} and SWAP-NAS~\cite{peng2024swapnassamplewiseactivationpatterns} struck a promising balance between search efficiency and performance. Nevertheless, these methods still face concerns regarding lack of theoretical guarantees and the disparity between proxy metrics and actual performance.
\begin{figure}[t]
    \includegraphics[width=1\textwidth]{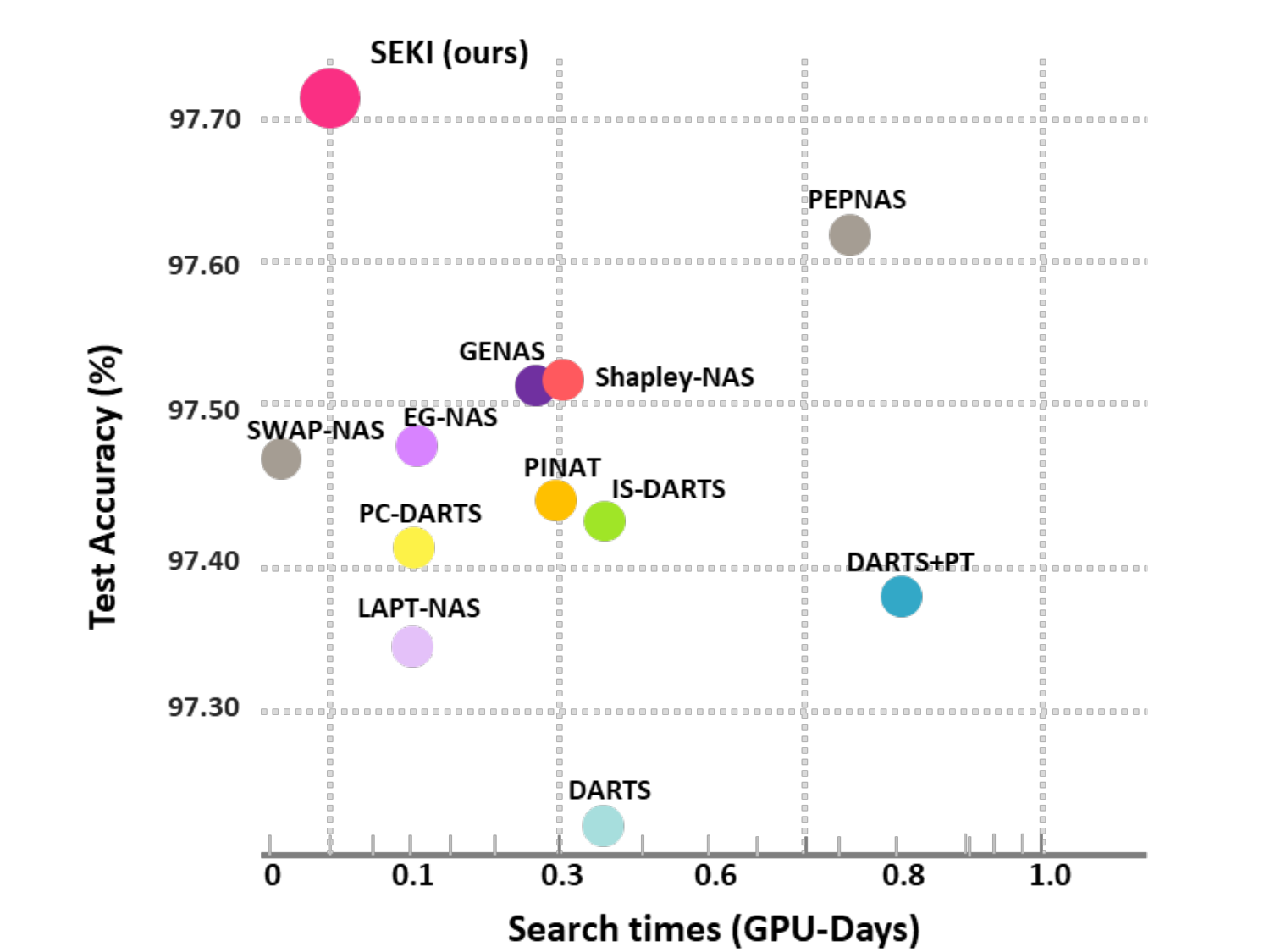}
  \caption{Speed-performance comparison of our proposed SEKI with other NAS methods on CIFAR-10 (methods over 1 GPU day are not included).}
  \label{fig:acc}
  \vspace{-4mm}
\end{figure}

Recently, with the growing capabilities of large language models (LLMs), several studies have started exploring the use of LLMs for NAS~\cite{chen2023evopromptinglanguagemodelscodelevel,zhang2023automlgptautomaticmachinelearning,nasir2024llmatic,qin2024flnasfairnessnasresource,wang2024graphneuralarchitecturesearch,dong2023heterogeneous}. By enabling LLMs to output design principles, these methods significantly enhances interpretability, offering insights into the architecture design process. However, most of these studies are still in the early exploratory stages and have yet to achieve competitive results. LAPT-NAS~\cite{zhou2024designprincipletransferneural} is one of the most successful recent works in this area. It introduced an innovative approach that uses LLMs to learn design principles from existing neural architectures and transfer them to new tasks, achieving both high search efficiency and competitive performance. 
However, 
LAPT-NAS heavily depends on a vast amount of relevant data from existing architectures to establish its design principles. This reliance presents a significant challenge for researchers, especially in domains where such data is scarce. Additionally, it exhibits limitations in dynamic optimization and in exploring entirely new architectural directions. If the quality of the initial principles is insufficient, it can greatly affect subsequent search and optimization processes. Consequently, LAPT-NAS still falls short of surpassing earlier non-LLM-based methods. Overall, we believe that the potential of LLMs in NAS research remains far from fully explored.
\begin{figure}[ht]
  \centering 
    \includegraphics[width=1.0\textwidth]{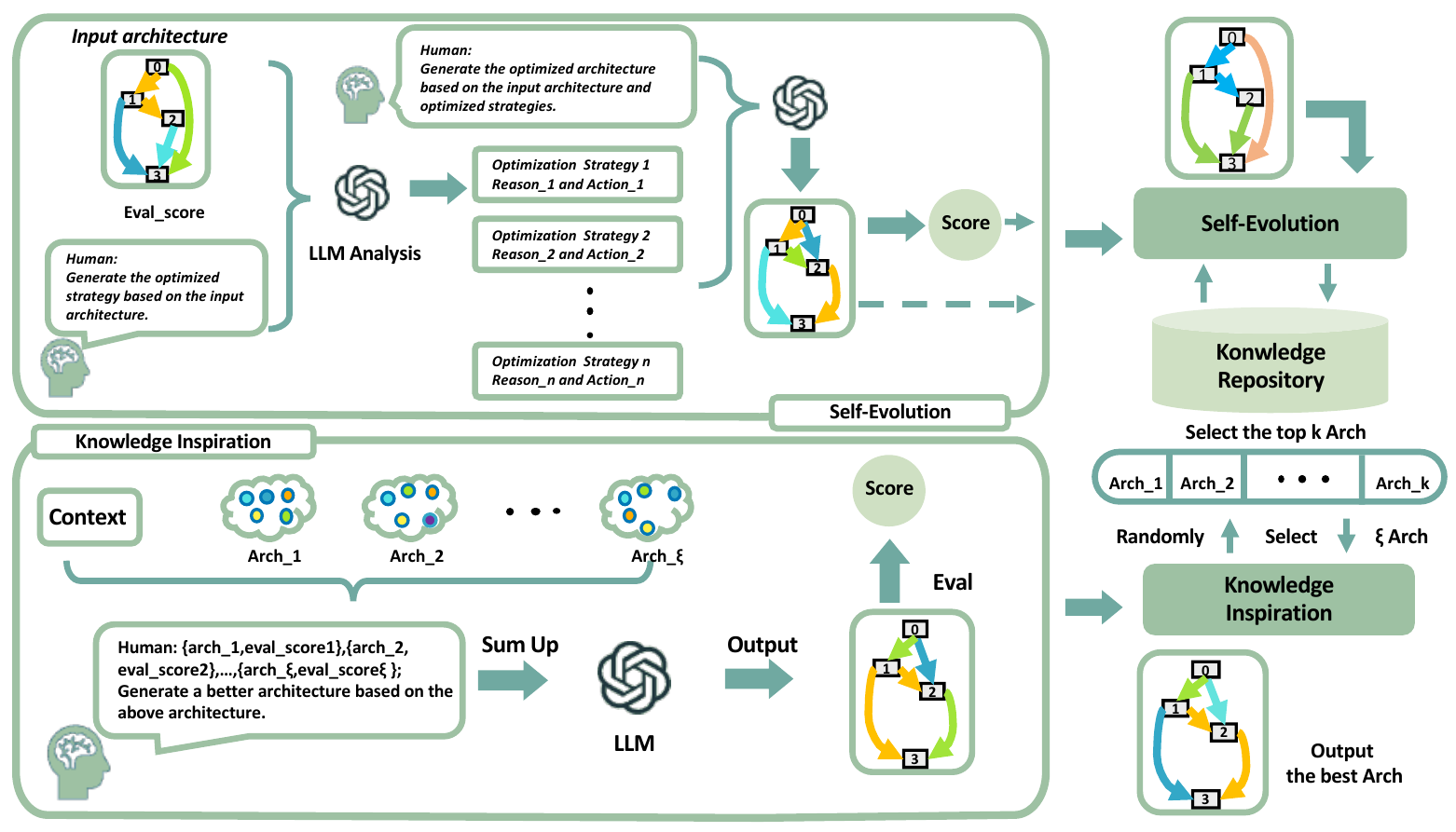}
  \caption{Framework of SEKI. SEKI is composed of two stages: self-evolution and knowledge inspiration. In each iteration of the self-evolution, the LLM generates optimization strategies and produces a new, refined architecture by analyzing the current architecture and its performance metrics. Over successive iterations, we compile a collection of high-performing architectures throughout this process and store the top $k$ architectures in a knowledge repository. Then in knowledge inspiration, by summarizing and analyzing $\xi$ validated high-quality architectures from knowledge repository, the LLM extracts common design patterns and generates new candidate architectures.}
  \label{fig:all}
  \vspace{-3mm}
\end{figure}

In this paper, we present a novel LLM-based NAS solution, {\bf SEKI} (Self-Evolution and Knowledge Inspiration from LLMs), designed to more effectively explore the potential of LLMs in enhancing the efficiency and performance of architecture search, without relying on any data on existing architectures. Due to the absence of data, the LLM initially lacks sufficient reference examples, making it challenging to directly generate high-performance architectures. To overcome this limitation, we introduce a {\bf self-evolution process} as the first stage. In each iteration, the LLM generates optimization strategies and produces a new, refined architecture by analyzing the current architecture and its performance metrics. Over successive iterations, the quality of the generated architectures progressively improves. We compile a collection of high-performing architectures throughout this process, storing the top $k$ architectures in a {\bf knowledge repository}. In the second stage, we leverage this repository for {\bf knowledge inspiration}. By summarizing and analyzing $\xi$ ($\xi < k$) validated high-quality architectures, the LLM extracts common design patterns and directly generates new candidate architectures. New candidates are added to knowledge repository for furture iterations. SEKI seamlessly integrates self-evolution and knowledge inspiration through LLMs, demonstrating strong capabilities in both dynamic optimization and the exploration of new architectural directions. As a result, it provides an efficient, reliable, and practical solution for neural architecture design. 

To the best of our knowledge, SEKI is the first LLM-based solution that achieves state-of-the-art (SOTA) results across various datasets and search spaces: it requires just 0.05 GPU-days for the search process and achieves 97.71$\%$ and 84.14$\%$ top-1 accuracy on CIFAR-10 and CIFAR-100, respectively 
(\Cref{table:cifar10}), and 75.8$\%$ on ImageNet. Furthermore, a direct search on ImageNet yields a top-1 accuracy of 76.1$\%$ at a cost of only 2.0 GPU-days on a single RTX A100 GPU (\Cref{tab:image}), outperforming existing SOTA methods in both efficiency and performance. It also achieves SOTA competitive results across various tasks (\Cref{table:ab2}), demonstrating a strong generalization capability. While SEKI is simple, it shows the success and potentials of LLM in NAS.

\section{Related work}

Neural Architecture Search (NAS) is a technique designed to automate the creation of neural network architectures. Its primary goal is to enhance model performance and simplify the design process by algorithmically identifying optimal network structures. Traditional NAS methods~\cite{2017Learning, 2018Regularized} often involve training and evaluating a large number of candidate architectures from scratch, making them both computationally expensive and time-consuming.
To address these challenges and improve efficiency, researchers have proposed several advancements in recent years. Among these, {\bf One-Shot NAS} and the integration of {\bf LLMs for NAS} have emerged as two prominent and promising directions.

\subsection{One-Shot NAS}
One-Shot NAS \cite{brock2017smashoneshotmodelarchitecture} has revolutionized NAS by introducing the concept of a supernet, enabling the performance evaluation of multiple architectures without the need for individually training them. Early works such as \cite{2018Efficient,liu2018darts,xu2019pc} laid the groundwork for this approach, demonstrating the effectiveness of weight-sharing techniques. However, these methods faced several challenges, including weight entanglement and suboptimal fairness in architecture evaluation \cite{chen2019progressive,zela2019understanding}. Recent research has sought to address these limitations through various innovations. For instance, \cite{hu2020angle} introduced an angle-based metric to simplify the search space by pruning unpromising candidates, thereby reducing the difficulty of identifying high-quality architectures. Similarly, \cite{DBLP:journals/corr/abs-2006-10355} employed incremental learning to bridge the gap between the search and evaluation phases. In \cite{DBLP:journals/corr/abs-2108-11014}, node normalization and decorrelation discretization strategies were proposed to improve generality and stability. Additionally, \cite{xiao2022shapley} utilized the Shapley value to assess the importance of operations, while \cite{Cai_Chen_Liu_Ling_Lai_2024} introduced a hybrid approach that combined evolutionary strategies with gradient descent, effectively mitigating local issues through global optimization while maintaining efficiency. Despite these advancements, One-Shot NAS methods continue to struggle with several key challenges. They are particularly prone to local optima, especially in complex search spaces. Moreover, gradient-based One-Shot NAS approaches are highly sensitive to initial conditions and often encounter difficulties in thoroughly exploring non-smooth or multi-modal search spaces. Additionally, 
most existing NAS methods still depend on expert knowledge to design the search space, search algorithms, and performance evaluation systems \cite{chen2023evopromptinglanguagemodelscodelevel}. These limitations highlight the need for further innovation to fully unlock the potential of One-Shot NAS.

\begin{figure}[ht]
  \centering 
    \includegraphics[width=0.95\textwidth]{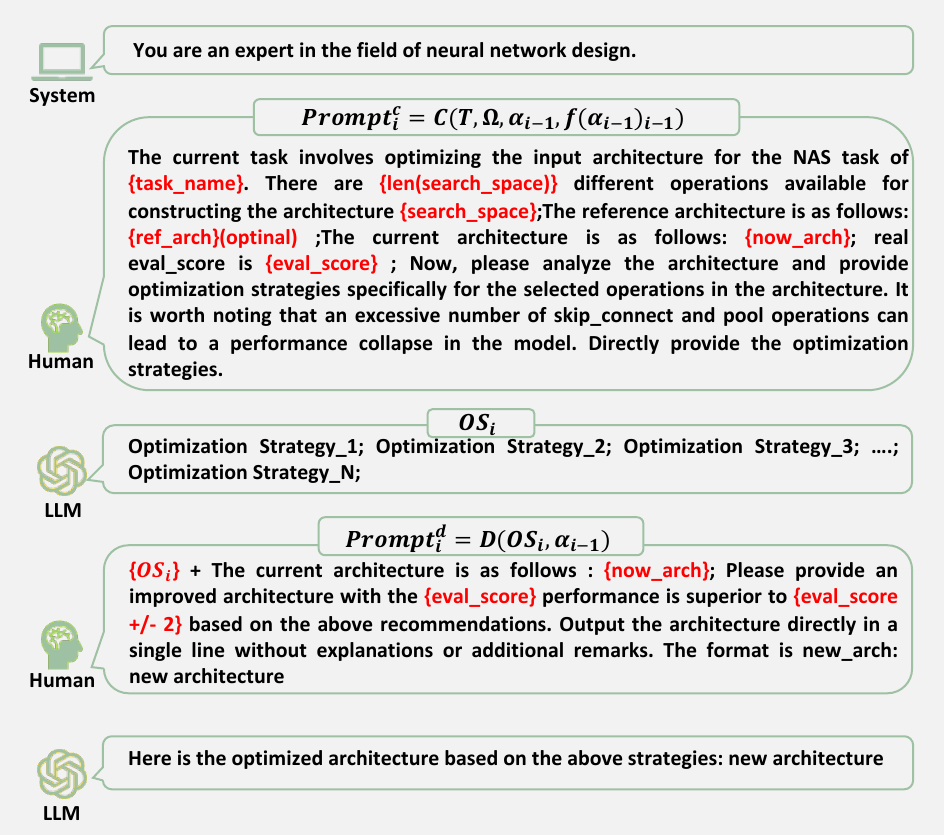}
  \caption{Prompt framework for Self-Evolution.}
  \label{fig:p1}
  \vspace{-4mm}
\end{figure}

\begin{figure}[ht]
  \centering 
    \includegraphics[width=0.95\textwidth]{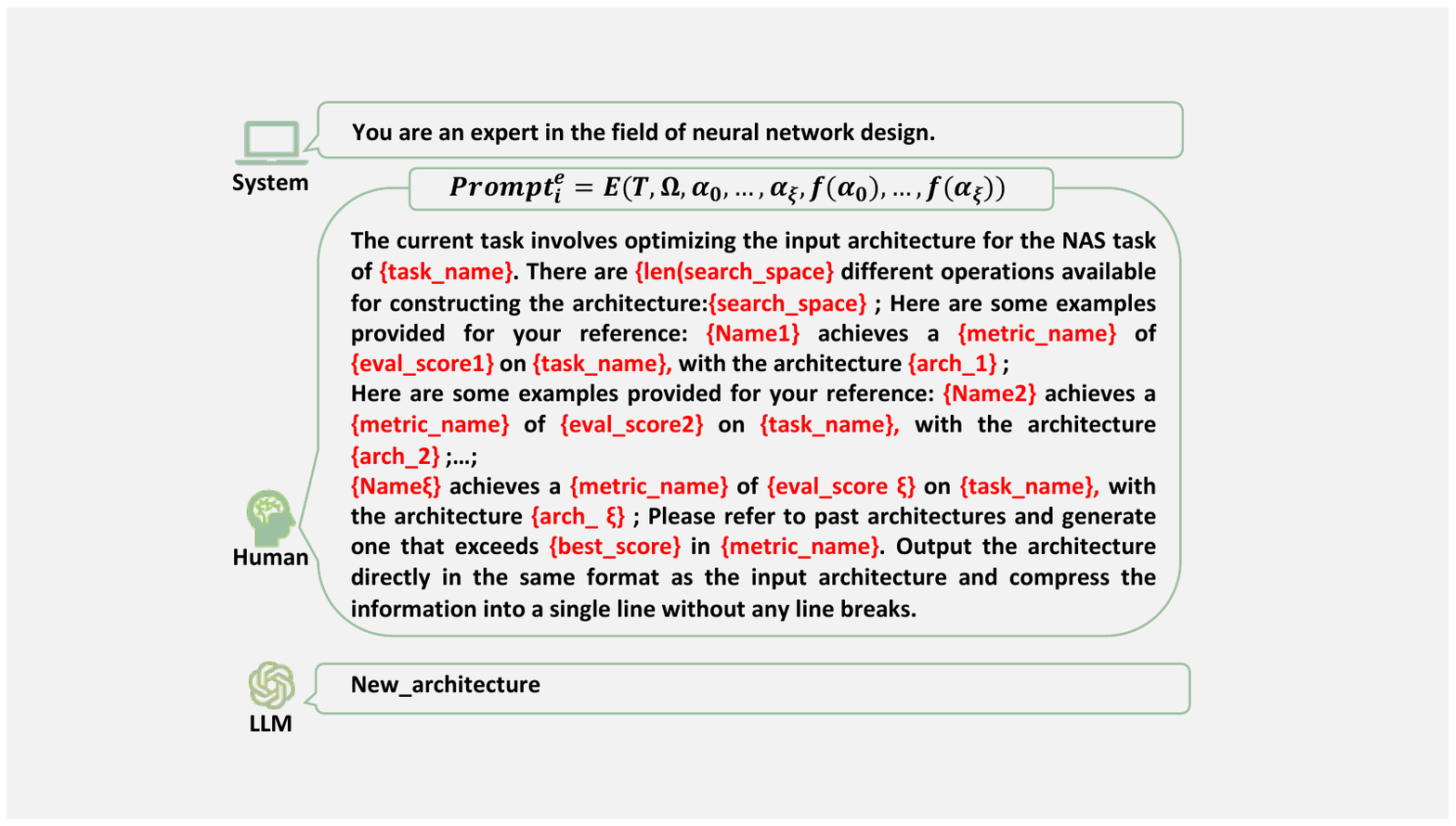}
  \caption{Prompt framework for Knowledge Inspiration.}
  \label{fig:p2}
  \vspace{-4mm}
\end{figure}

\subsection{LLM for NAS}

Recent advances have introduced LLMs into the NAS field, aiming to enhance the automation of architecture design. With their exceptional performance across various domain-specific tasks, LLMs have unlocked new opportunities in this area. For example, studies such as \cite{chen2023evopromptinglanguagemodelscodelevel,zhang2023automlgptautomaticmachinelearning,nasir2024llmatic,qin2024flnasfairnessnasresource,wang2024graphneuralarchitecturesearch,dong2023heterogeneous} have successfully employed LLMs, including GPT-4, to generate architectures for Convolutional Neural Networks (CNNs) and Graph Neural Networks (GNNs), achieving promising results. Beyond architecture generation, LLMs have been utilized as performance predictors to accelerate the NAS search process \cite{zhang2023automlgptautomaticmachinelearning,jawahar2024llmperformancepredictorsgood,chen2024large,10448075}. They have also been applied to optimize search spaces and design architectures, significantly improving the automation and interpretability of the overall process \cite{zhou2024designprincipletransferneural}. Despite their promising capabilities, the use of LLMs for NAS is still relatively new and comes with certain limitations, such as lower performance or a reliance on relevant data to achieve high-quality results. We believe the issue lies in the fact that the capabilities of LLMs have yet to be fully explored. To address these challenges, our solution, SEKI, fully harnesses the potential of LLMs through self-evolution and knowledge inspiration, and delivers both high efficiency and performance without the need for any data/ prior knowledge on architectures.

\section{Methods}
\section{Technical Approach}
We propose SEKI ({\bf S}elf-{\bf E}volution and {\bf K}nowledge {\bf I}nspiration from LLMs), a novel solution to effectively explore the potential of LLMs in enhancing NAS, without relying on any data or prior knowledge of the network. SEKI is composed of two stages: {\it Self-Evolution} and {\it Knowledge Inspiration} (\Cref{fig:all}). 
To begin with, the Self-Evolution method is used to optimize the architecture through step-by-step performance feedback. The optimal architectures and their corresponding evaluation results are stored in a knowledge repository, providing important reference for subsequent searches. 
As the architecture search progresses, the Knowledge Inspiration method begins to take effect. By analyzing the accumulated architecture and evaluation data in the knowledge repository, the LLM can draw valuable insights from historical knowledge to offer more precise optimization strategies for new architecture designs. The new designs are added to knowledge repository for future iterations. The guidance from knowledge not only helps identify potential optimization directions but also promotes the diversity and innovation of architecture designs.
The combination of these methods ensures that the architecture search is efficient 
while providing stronger robustness, 
facilitating better architecture optimization. 
Below we introduce self-evoluation and knowledge inspiration in details. 

\floatname{algorithm}{Algorithm}
\renewcommand{\algorithmicrequire}{\textbf{Require:}}
    \begin{algorithm}[htb]
    \setlength{\tabcolsep}{4pt}
        \caption{\textbf{Main framework of SEKI}}
        \label{al:ifenas}
        \begin{algorithmic}[1] 
            \Require{Search Space: $\boldsymbol{\Omega}$; a pre-trained LLM; Epoch $n$; Target task $T$, Architecture $\boldsymbol{\alpha}$, Knowledge Repository $S$, Sample times $\lambda$ and $\gamma$, with $\lambda + \gamma = n$.}
        \State Input the initialized architecture $\boldsymbol{\alpha}_{0}$ and calculate its evaluation score $f(\boldsymbol{\alpha}_{0})$;
        \For{$i=1,2,\ldots,n$}
            \While{$i < \lambda$}{}
                 \State Construct $\mathbf{Prompt}_{i}^{c}$ and $OS_{i}$ by Equ. \ref{eq-q1} and Equ. \ref{eq-os};
                 \State $\mathbf{Prompt}_{i}^{d}$ is derived from $\boldsymbol{\alpha}_{i-1}$ and $OS_{i}$ in Equ. \ref{eq-pp};
                 \State New architecture $\boldsymbol{\alpha}_{i}$  is searched by Equ. \ref{eq-qq};
                 \State Calculate the evaluation score $f(\boldsymbol{\alpha}_{i})$; 
                 \State Store both $\boldsymbol{\alpha}_{i}$ and $f(\boldsymbol{\alpha}_{i})$ in the knowledge repository $S = \left\{ \boldsymbol{\alpha}_{0}, \boldsymbol{\alpha}_{1},\ldots,\boldsymbol{\alpha}_{\lambda}, f(\boldsymbol{\alpha}_{0}), f(\boldsymbol{\alpha}_{1}),,\ldots,f(\boldsymbol{\alpha}_{\lambda}) \right\}$;
             \EndWhile
             \While{$i >= \lambda$}{}
                \State Select the top $k$ architectures from $S$, and further choose $\xi$ architectures from them to construct $\mathbf{Prompt}_{i}^{e}$ with a prompt template function $E(\cdot)$;
                \State $\boldsymbol{\alpha}_{i} \leftarrow LLM(\mathbf{Prompt}_{i}^{e})$;
                \State Calculate the evaluation score $f(\boldsymbol{\alpha}_{i})$; 
                \State Store both $\boldsymbol{\alpha}_{i}$ and $f(\boldsymbol{\alpha}_{i})$ in the knowledge repository $S = \left\{ \boldsymbol{\alpha}_{0},\boldsymbol{\alpha}_{1},\ldots,\boldsymbol{\alpha}_{n}, f(\boldsymbol{\alpha}_{0}),f(\boldsymbol{\alpha}_{1}),\ldots,f(\boldsymbol{\alpha}_{n}) \right\}$;
            \EndWhile
        \EndFor
    \State Output the best architecture $\boldsymbol{\alpha}_{best}$ from $S$.
        \end{algorithmic}
    \end{algorithm}
   \floatname{algorithm}{Algorithm}
   \vspace{-2mm}

\subsection{Self-Evolution}
Inspired from the chain-of-thought approach in LLM, the main idea of Self-Evolution is to break down a complex problem into multiple subproblems, which LLMs can then solve step by step, thereby gradually addressing the overall problem and achieving better performance in specific tasks. Specifically, we run self-evolution for $\lambda$ rounds. In each round $i$, there are four steps: 
\vspace{-1mm}
\begin{itemize}
\item[$\bullet$] 
{\bf Step 1: }
We construct $\mathbf{Prompt}_{i}^{c}$ based on the target task $T$, search space $\boldsymbol{\Omega}$, input architecture $\boldsymbol{\alpha}_{i-1}$, and its evaluation score $f(\boldsymbol{\alpha}_{i-1})$ using the prompt template $C(\cdot)$. An example of $C(\cdot)$ is in~\Cref{fig:p1}. More details can be referred to Appendix. 
\vspace{-1mm}
\begin{equation}
\label{eq-q1}
\mathbf{Prompt}_{i}^{c} = C(T,\boldsymbol{\Omega},\boldsymbol{\alpha}_{i-1},f(\boldsymbol{\alpha}_{i-1}))
\end{equation}
\item[$\bullet$]
\vspace{-1mm}
{ \bf Step 2: }
Given the created prompt, the LLM analyzes the potential issues of the input architecture $\boldsymbol{\alpha}_{i-1}$, evaluates possible optimization directions, and generates corresponding optimization strategies $OS_{i}$:
\begin{equation}
\label{eq-os}
OS_{i} \leftarrow LLM(\mathbf{Prompt}_{i}^{c}) 
\end{equation}
\item[$\bullet$]
\vspace{-1mm}
{\bf Step 3:}
A new prompt $\mathbf{Prompt}_{i}^{d}$ is constructed with the optimization strategy $OS_{i}$ and architecture $\boldsymbol{\alpha}_{i-1}$ by the template function $D(\cdot)$. An example of $D(\cdot)$ is in~\Cref{fig:p1}. More details can be referred to Appendix.
\vspace{-1mm}
\begin{equation}
\label{eq-pp}
\mathbf{Prompt}_{i}^{d} = D(OS_{i},\boldsymbol{\alpha}_{i-1})
\end{equation}
\item[$\bullet$] 
\vspace{-1mm}
{\bf Step 4:}
$\mathbf{Prompt}_{i}^{d}$ is provided to the LLM, which generates the new architecture $\boldsymbol{\alpha}_{i}$: 
\vspace{-1mm}
\begin{equation}
\label{eq-qq}
\boldsymbol{\alpha}_{i} \leftarrow LLM(\mathbf{Prompt}_{i}^{d})
\end{equation}
\end{itemize}
\vspace{-1mm}
Unlike traditional methods that require a large amount of downstream task data for supervised fine-tuning before directly generating optimal solutions, Self-Evolution leverages the general knowledge and learning abilities of LLMs. By gradually decomposing the problem and performing targeted optimization, LLMs can efficiently optimize the architecture without the need for extensive data support. 


\begin{table*}[!htb]
\centering
\caption{Comparison of SEKI with SOTA image classifiers on CIFAR-10 and CIFAR-100, on DARTS search space. The results of SEKI are obtained from repeated experiments with 4 random seeds. Best result is marked with \textbf{bold} and the second best result is marked with \underline{underline}.}
\scriptsize
\begin{tabular}{lcccccc}
\hline
\multirow{2}{*}{Architecture} & \multicolumn{2}{c}{Test Error, top-1 (\%)} & Params & Search Cost & \multicolumn{1}{l}{\multirow{2}{*}{Search Method}} & \multicolumn{1}{l}{\multirow{2}{*}{Evaluation}} \\ \cline{2-3}
 & CIFAR-10 & CIFAR-100 & (M) & (GPU-Days) & \multicolumn{1}{l}{} & \multicolumn{1}{l}{} \\ \hline
ResNet \cite{2016Deep} & 4.61 & 22.1 & 1.7 & - & - & - \\ \hline
ENAS + cutout \cite{2018Efficient} & 2.89 & - & 4.6 & 0.5 & Reinforce & One-shot \\
AmoebaNet-A \cite{2018Regularized} & 3.34 & 17.63 & 3.3 & 3150 & Evolution & - \\
NSGA-Net \cite{DBLP:journals/corr/abs-1810-03522} & 2.75 & 20.74 & 3.3 & 4.0 & Evolution & - \\
NSGANetV1-A2 \cite{DBLP:journals/corr/abs-1810-03522} & 2.65 & - & 0.9 & 27 & Evolution & - \\
EPCNAS-C \cite{9930866} & 3.24 & 18.36 & 1.44 & 1.2 & Evolution & One-Shot \\
EAEPSO \cite{10045029} & 2.74 & 16.94 & 2.94 & 2.2 & Evolution & One-Shot \\
PINAT\cite{lu2023pinat} & 2.54 & - & 3.6 & 0.3 & Evolution & Predictor \\
SWAP-NAS \cite{peng2024swapnassamplewiseactivationpatterns} & 2.54 & - & 3.48 & \bf{0.004} & Evolution & Training-free \\
PEPNAS \cite{10508441} & 2.38 & 16.46 & 4.23 & 0.7 & Evolution & - \\ 
DARTS(1st) \cite{liu2018darts} & 3.00 & 17.54 & 3.4 & 0.4 & Gradient & One-Shot \\
ProxylessNAS + cutout \cite{cai2018proxylessnas} & \textbf{2.02} & - & 5.7 & 4.0 & Gradient & One-Shot \\
PC-DARTS + cutout \cite{xu2019pc} & 2.57 & 16.90 & 3.6 & 0.1 & Gradient & One-Shot \\
Fair-DARTS \cite{DBLP:journals/corr/abs-1911-12126} & 2.54 & - & 2.8 & 0.4 & Gradient & One-Shot \\
BayesNAS \cite{DBLP:journals/corr/abs-1905-04919} & 2.81 & - & 3.4 & 0.2 & Gradient & One-Shot \\
MiLeNAS + cutout \cite{he2020milenas} & 2.76 & - & 2.09 & 0.3 & Gradient & One-Shot \\
DARTS+PT \cite{wang2021rethinking} & 2.61 & - & 3.0 & 0.8 & Gradient & One-Shot \\
$\beta$-DARTS + cutout\cite{ye2022beta} & 2.53 & 16.24 & 3.75/3.80 & 0.4 & Gradient & One-Shot \\
Shapley-NAS + cutout \cite{xiao2022shapley} & 2.47 & - & 3.4 & 0.3 & MCMC & One-Shot \\
IS-DARTS \cite{he2023isdartsstabilizingdartsprecise} & 2.56 & - & 4.25 & 0.4 & Gradient & One-Shot \\
LAPT-NAS \cite{zhou2024designprincipletransferneural} & 2.65 & - & - & 0.1 & LLM & - \\
EG-NAS \cite{Cai_Chen_Liu_Ling_Lai_2024} & 2.53 & \underline{16.22} & 3.2 & 0.1 & Mixed & One-Shot \\
GENAS \cite{xue2024gradient} & 2.49 & 16.96 & 3.2 & 0.26 & Evolution & One-Shot \\  \hline
SEKI & \underline{2.29} & \textbf{15.86} & 3.92 & \underline{0.05} & LLM & One-Shot \\
\hline
\end{tabular}
\label{table:cifar10}
\vspace{-3mm}
\end{table*}

\subsection{Knowledge Inspiration}
LLMs can effectively leverage past dialogue history, extracting key insights to deliver precise responses to user queries. This ability not only showcases their powerful learning and knowledge-driven capabilities but also underscores how their extensive knowledge repository enhances answer accuracy. 

In NAS, following this principle, we first collect extensive architectural and performance evaluation data from the Self-Evolution stage to construct a knowledge repository. We then introduce the Knowledge Inspiration method, which analyzes and leverages this accumulated knowledge to generate new and enhanced architectures.
Specifically in each iteration of Knowledge Inspiration $i$, we first select the top $k$ solutions from the knowledge repository, and then randomly choose $\xi$ ($<k$) solutions from them to use in new architecture search. The core idea behind this approach is to introduce randomness, reducing over-reliance on optimal solutions. This helps prevent the search from becoming trapped in local optima thus excessively generating repetitive architectures. As a result, the diversity of the architecture search are broadened. Then we create a prompt $\mathbf{Prompt}_{i}^{e}$ using prompt template $E(\cdot)$ shown in ~\Cref{fig:p2} (more details can be referred to Appendix), and input the prompt to LLM to search for new architecture $\boldsymbol{\alpha}_{i}$. The newly discovered architectures $\boldsymbol{\alpha}_{i}$ are continuously added to the knowledge repository, allowing the optimization strategy to be refined over time.

The detailed algorithm of SEKI is in \Cref{al:ifenas}.

\section{Experiments}

In this section, we compare SEKI with other well-known NAS methods. Specifically, we conduct experiments on diverse tasks across different architecture search spaces. We begin by introducing the search spaces and experimental setup, followed by presenting the results and comparisons.  


\begin{table}[htb]
\centering
\footnotesize
\begin{threeparttable}
\caption{Comparison with SOTA image classifiers on ImageNet-1K. $\dagger$ indicates results obtained by searching on ImageNet; otherwise, the search is conducted on CIFAR-10. All searches are performed within the DARTS search space. Best result is marked with \textbf{bold}.}
\begin{tabular}{lccc}
\hline
\multirow{2}{*}{\footnotesize Architecture} & Test error & Search cost & Params \\
 & top-1(\%) & (GPU-Days) & (M) \\ \hline
Inception-v1 & 30.1 & - & 6.6 \\
MobileNet & 29.4 & - & 4.2 \\ \hline
NASNet-A & 26.0 & 2000 & 3.3 \\
NASNet-B & 27.2 & 2000 & 3.3 \\
NASNet-C & 27.5 & 2000 & 3.3 \\
AmoebaNet-A & 25.5 & 3150 & 3.2 \\
NSGANetV1-A2 & 25.5 & 27 & 4.1 \\ 
EAEPSO & 26.9 & 4.0 & 4.9 \\
EPCNAS-C2 & 27.1 & 1.17 & 3.0 \\
DARTS(2st) & 26.7 & 1.0 & 4.7 \\
SNAS & 27.3 & 1.5 & 2.8 \\
ProxylessNAS$^{\dagger}$ & 24.9  & 8.3    & 7.1   \\ 
GDAS & 26.0 & 0.3 & 3.4 \\
BayesNAS & 26.5 & 0.2 & 3.9\\
PC-DARTS & 25.1 & 0.1 & 4.7 \\ 
DrNAS$^{\dagger}$& 24.2 & 4.6 & 5.7 \\
DARTS+PT$^{\dagger}$&25.5&3.4&4.7 \\
G-NAS &27.6 &3.8&14.4 \\
PINAT& 24.9 & 0.3 & 5.2 \\
EG-NAS& 24.9 & 0.1 & 5.3 \\
LAPT-NAS& 24.9 & 2.0 & 4.6 \\ \hline
SEKI & 24.5 & \bf{0.05} & 5.2 \\
SEKI$^{\dagger}$ & \bf{23.9} & 2.0 & 5.5 \\ \hline 
\end{tabular}
\label{tab:image}
\end{threeparttable}
\vspace{-3mm}
\end{table}

\setlength{\tabcolsep}{4pt}
\begin{table*}[!htb]
\centering
\caption{Performance comparison on NAS201. Note that SEKI is searched only on the CIFAR-10 dataset yet achieves competitive results on CIFAR-10, CIFAR-100, and ImageNet16-120. The reported average values are based on four independent search runs. Best result is marked with \textbf{bold} and the second best result is marked with \underline{underline}. Top-1 accuracy is measured.}
\scriptsize
\begin{tabular}{lcccccc}
\hline
\multirow{2}{*}{\footnotesize{Architecture}} & \multicolumn{2}{c}{CIFAR-10} & \multicolumn{2}{c}{CIFAR-100} & \multicolumn{2}{c}{ImageNet16-120} \\ \cline{2-7} 
 & valid & test & valid & test & valid & test \\ \hline
ResNet\cite{2016Deep} & 90.83 & 93.97 & 70.42 & 70.86 & 44.53 & 43.63 \\ \hline
Random (baseline) & 90.93$\pm$0.36 & 93.70$\pm$0.36 & 70.60$\pm$1.37 & 70.65$\pm$1.38 & 42.92$\pm$2.00 & 42.96$\pm$2.15 \\
ENAS \cite{2018Efficient}& 37.51$\pm$3.19 & 53.89$\pm$0.58 & 13.37$\pm$2.35 & 13.96$\pm$2.33 & 15.06$\pm$1.95 & 14.57$\pm$2.10 \\
RandomNAS \cite{2019Random}& 80.42$\pm$3.58 & 84.07$\pm$3.61 & 52.12$\pm$5.55 & 52.31$\pm$5.77 & 27.22$\pm$3.24 & 26.28$\pm$3.09 \\
SETN \cite{dong2019one} & 84.04$\pm$0.28 & 87.64$\pm$0.00 & 58.86$\pm$0.06 & 59.05$\pm$0.24 & 33.06$\pm$0.02 & 32.52$\pm$0.21 \\ 
GDAS \cite{2019Searching}  & 90.01$\pm$0.46 & 93.23$\pm$0.23 & 24.05$\pm$8.12 & 24.20$\pm$8.08 & 40.66$\pm$0.00 & 41.02$\pm$0.00 \\
DSNAS \cite{2020DSNAS} & 89.66$ \pm$0.29 & 93.08±0.13 & 30.87$\pm$16.40 & 31.01$\pm$16.38 & 40.61$\pm$0.09 & 41.07$\pm$0.09 \\
DARTS (2st) \cite{liu2018darts} & 39.77 & 54.30 & 15.03 & 15.61 & 16.43 & 16.32 \\
PC-DARTS \cite{xu2019pc} & 89.96$\pm$0.15 & 93.41$\pm$0.30 & 67.12$\pm$0.39 & 67.48$\pm$0.89 & 40.83$\pm$0.08 & 41.31$\pm$0.22 \\
iDARTS \cite{DBLP:journals/corr/abs-2108-11014} & 89.86$\pm$0.60 & 93.58$\pm$0.32 & 70.57$\pm$0.24 & 70.83$\pm$0.48 & 40.38$\pm$0.59 & 40.89$\pm$0.68 \\ 
DARTS- \cite{DBLP:journals/corr/abs-2009-01027}& 91.03$\pm$0.44 & 93.80$\pm$0.40 & 71.36$\pm$1.51 & 71.53$\pm$1.51 & 44.87$\pm$1.46 & 45.12$\pm$0.82\\
IS-DARTS \cite{he2023isdartsstabilizingdartsprecise} & \bf{91.55} & \bf{94.36} & \bf{73.49} & \bf{73.51} & \underline{46.37} & \underline{46.34} \\ 
LLMatic\cite{nasir2024llmatic}& - & 94.26$\pm$0.13 & - & 71.62$\pm$1.73 & - & 45.87$\pm$0.96 \\
LEMO-NADE\cite{rahman2024lemonademultiparameterneuralarchitecture}& 90.90 & 89.41 & 68.38 & 67.90 & 27.05 & 27.70 \\
EG-NAS \cite{Cai_Chen_Liu_Ling_Lai_2024} & 90.12$\pm$0.05 & 93.56$\pm$0.02 & 70.78$\pm$0.12 & 70.91$\pm$0.07 & 44.89$\pm$0.29 & 46.13$\pm$0.46 \\
\hline
SEKI & \underline{91.44$\pm$1.26} & \underline{94.34$\pm$0.35} & \underline{72.74$\pm$1.92} & \underline{72.75$\pm$0.37} & \textbf{46.56$\pm$0.56} & \textbf{46.90$\pm$0.19} \\ \hline
Oracle Best & 91.61 & 94.37 & 73.49 & 73.51 & 46.77 & 47.31 \\ \hline
\end{tabular}
\label{table:nasbench201}
\end{table*}
 
\begin{table*}[ht]
\centering
\footnotesize
\caption{Results on Trans101 (For metrics: $\uparrow$ indicates that higher values are better, while $\downarrow$ indicates that lower values are better). SEKI is evaluated over five runs with different random seeds. Best result is marked with \textbf{bold} and the second best result is marked with \underline{underline}.}
\begin{tabular}{l|cccccccc}
\hline
Tasks & Cls.O. & Cls.S. & Auto. & Normal & Sem.Seg. & Room. & \multicolumn{1}{c|}{Jigsaw} & Total \\ \hline
Metric & Acc$\uparrow$ & Acc$\uparrow$ & SSIM$\uparrow$ & SSIM$\uparrow$ & mIoU$\uparrow$ & L2loss $\downarrow$ & \multicolumn{1}{c|}{Acc $\uparrow$} & Ave. Rank$\downarrow$ \\ \hline
RS & 45.16 & 54.41 & 55.94 & 56.85 & 25.21 & 61.48 & 94.47 & 85.61 \\
REA & 45.39 & 54.62 & \underline{56.96} & 57.22 & 25.52 & 61.75 & 94.62 & 38.50 \\
BONAS & 45.50 & 54.46 & 56.73 & 57.46 & 25.32 & 61.10 & \underline{94.81} & 34.31 \\
weakNAS-t & \bf{47.40} & 54.78 & 56.90 & 57.19 & 25.41 & 60.70 & - & 35.73 \\
Arch-Graph-zero & 45.64 & \underline{54.80} & 56.61 & 57.90 & 25.73 & 60.21 & - & 14.7 \\
Arch-Graph & 45.81 & \textbf{54.90} & 56.58 & \textbf{58.27} & \textbf{26.27} & \underline{59.38} & - & \underline{12.2} \\
LAPT-NAS & 45.96 & - & 56.52 & 57.69 & \underline{25.91} & 60.18 & - & 12.3 \\ \hline
SEKI & \underline{46.22 } & 54.63  & \textbf{57.03} & \underline{58.22}  & 25.80  & \textbf{59.37 } & \textbf{94.99 } & \textbf{7.7} \\ \hline
Oracle Best & 46.32 & 54.93 & 57.72 & 59.62 & 26.27 & 59.37 & 95.37 & 1 \\ \hline
\end{tabular}
\label{table:ab2}
\end{table*}

\begin{table}[ht]
\centering
\footnotesize
                                     \caption{Comparison of Self-Evolution and Knowledge Inspiration Coefficients $\lambda$ and $\gamma$ on SEKI. 
                                     }
\begin{tabular}{cccc}
\hline
\multicolumn{1}{l|}{$\lambda$} & \begin{tabular}[c]{@{}c@{}}Top-1 Acc\\ (\%)\end{tabular} & \begin{tabular}[c]{@{}c@{}}Params\\ (M)\end{tabular} & \begin{tabular}[c]{@{}c@{}}Search Cost\\ (min)\end{tabular} \\ \hline
15 & 85.15 & 3.62 & 15 \\
25 & 85.54 & 3.16 & 16 \\
30 & 85.74 & 3.40 & 18 \\ 
35 & \bf{86.33} & 3.77 & 19 \\
40 & 85.93 & 3.40 & 20 \\
45 & 85.35 & 3.57 & 24 \\ \hline
\end{tabular}
\label{table:ab1}
\vspace{-3mm}
\end{table}

\begin{table}[ht]
\centering
\footnotesize
\caption{Comparison of Knowledge Repository Variables $k$ and $\xi$ on SEKI. 
}
\begin{tabular}{l|cccc}
\hline
\multirow{2}{*}{$k$} & \multicolumn{2}{c|}{$\xi$ = $k$} & \multicolumn{2}{c}{$\xi$ = $3k/4$} \\ \cline{2-5} 
 & top-1\_acc(\%) & \multicolumn{1}{c|}{params(M)} & top-1\_acc(\%) & params(M) \\ \hline
8 & 85.54 & \multicolumn{1}{c|}{3.04} & 85.35 & 3.15 \\
16 & 85.35 & \multicolumn{1}{c|}{3.55} & 85.74 & 3.41 \\
24 & 84.57 & \multicolumn{1}{c|}{3.24} & 85.35 & 3.15 \\
32 & 84.17 & \multicolumn{1}{c|}{3.80} & 84.37 & 3.34 \\ \hline
\multirow{2}{*}{$k$} & \multicolumn{2}{c|}{$\xi$ = $k/2$} & \multicolumn{2}{c}{$\xi$ = $k/4$} \\ \cline{2-5} 
 & top-1\_acc(\%) & params(M) & top-1\_acc(\%) & params(M) \\ \hline
8 & 85.15 & \multicolumn{1}{c|}{3.78} & 83.59 & 3.12 \\
16 & \bf{86.33} & \multicolumn{1}{c|}{3.77} & 84.37 & 3.45 \\
24 & 85.74 & \multicolumn{1}{c|}{3.30} & 84.17 & 3.44 \\
32 & 85.35 & \multicolumn{1}{c|}{3.81} & 85.15 & 3.45 \\ \hline
\end{tabular}
\label{table:ab21}
\vspace{-3mm}
\end{table}

\subsection{Search Space and Experiment Setup}

We evaluate SEKI on three of the most widely used search spaces: DARTS~\cite{liu2018darts}, NAS201~\cite{DBLP:journals/corr/abs-2001-00326}, and Trans101~\cite{duan2021transnas}. These search spaces, along with their corresponding experimental setups, are introduced in details below. We use Qwen2.5-32B~\cite{qwen2.5} as the LLM in our experiments, as it is an open-source model that is accessible to researchers. We also run experiments with GPT4o-mini~\cite{openai2024gpt4ocard}, which yields similar results shown in \Cref{table:basemodel}. 

\subsubsection{DARTS}
In the DARTS search space, architectures adopt a cell-based structure, comprising two types of cells: normal cells and reduction cells. Each cell has two input layers and four sequential stages. Each stage consists of two layers and offers eight candidate operators to process information from the previous stage or input layers, resulting in approximately $10^9$ possible cell structures. Additionally, as both normal and reduction cells are optimized jointly, the total number of possible architectures reaches $(10^9)^2$ ~\cite{liu2018darts}.

We follow the literature to train the supernet on CIFAR-10 for 50 epochs with a batch size of 256. Then we perform 50 rounds of iterations to search the optimal network, i.e., $\lambda + \gamma = 50$, where $\lambda$ and $\gamma$ represent the number of iterations in the self-evolution and knowledge inspiration stages, respectively. Specifically, we set $\lambda =35 $ and $\gamma= 15$. During the knowledge inspiration stage, we use $k=16$ and $\xi=8$. We found these values to be optimal in \Cref{exp:as}.  
To evaluate the searched network architecture, we follow the literature to train it from scratch for 600 epochs on CIFAR-10 and CIFAR-100, respectively, with batch size at 96. For the ImageNet-1K dataset, the batch size is set to 1024, and the network is trained for 250 epochs. The remaining search and evaluation phases follow the same configuration as PC-DARTS~\cite{xu2019pc}.


\subsubsection{NAS201}
NAS-Bench-201 (NAS201) features architectures with repeated cells. Each cell consists of six layers, with five candidate operators available for each layer, resulting in a total of 15,625 distinct neural network architectures~\cite{DBLP:journals/corr/abs-2001-00326}.
 It offers a standardized testing environment for evaluating NAS algorithms on CIFAR-10, CIFAR-100, and ImageNet16-120 datasets.

To ensure a robust evaluation of the results, we derive the mean and standard deviation of the best architectures by conducting independent runs with four different random seeds. 
Similar to DARTS, the search is performed for 50 rounds, with $\lambda = 35$ and $\gamma = 15$, $k=16$ and $\xi=8$. The remaining setup follows $\beta$-DARTS~\cite{ye2022beta}.

\subsubsection{Trans101}
The architectures in TransNAS-Bench-101 (Trans101) follow the same cell-based structure as those in NAS201 but are distinguished by a more restricted set of candidate operators, consisting of only four options. This limitation reduces the search space to 4,000 possible candidate architectures~\cite{duan2021transnas}.

We perform architecture searches to tackle 7 diverse computer vision tasks: Object Classification (Obj), Scene Classification (SC), Room Layout (Roo), AutoEncoder (Auto), Jigsaw Puzzle (Jigsaw), Surface Normal (Nor), and Semantic Segmentation (Seg). Similar to the other two search spaces, the search process is conducted over 50 rounds with hyperparameters set to $\lambda = 35$ and $\gamma = 15$,             $k=16$ and $\xi=8$. To ensure robustness and reproducibility, the results on Trans101 are obtained through experiments using four different random seeds.

\subsection{Results and Comparison}          

In this section, we present results of SEKI on different search spaces and compare it to other well-known NAS methods.

\subsubsection{Results on DARTS}
 \Cref{table:cifar10} compares SEKI with other leading NAS methods on CIFAR-10 and CIFAR-100. SEKI achieves a test error rate of 2.29\% on CIFAR-10 with just 0.05 GPU-Days, significantly outperforming DARTS in both cost and accuracy. While ProxylessNAS achieves better accuracy and SWAP-NAS demonstrates higher search efficiency, SEKI delivers excellent performance while maintaining exceptionally low search costs, making it particularly well-suited for balancing cost and performance. Furthermore, the architectures discovered by SEKI on CIFAR-10 continue to excel when transferred to CIFAR-100, achieving a test error rate of 15.86\% and setting a new state-of-the-art.

\Cref{tab:image} compares SEKI to other methods on ImageNet. By transferring our optimized CIFAR-10 architecture to ImageNet, we achieved a top-1 test error rate of 24.5\%, demonstrating the strong generalization ability of our method. Furthermore, by performing a direct search on ImageNet (i.e., training the superior network directly on ImageNet), we achieved a top-1 test error rate of 23.9\% (a new SOTA) with an exceptionally low search cost of just 2.0 GPU-Days, underscoring the efficiency of SEKI. 
The searched final architectures can be found in Appendix. 

\subsubsection{Results on NAS201}

On NAS201, we compare SEKI with advanced NAS methods in  \Cref{table:nasbench201} across the CIFAR-10, CIFAR-100, and ImageNet16-120 datasets. SEKI achieves the second-best performance on both the CIFAR-10 and CIFAR-100 datasets, and the best performance on the ImageNet16-120 dataset, while being 8x more efficient (as shown in \Cref{table:cifar10}) than the comparable solution, IS-DARTS. SEKI is also very close to the oracle best.

\subsubsection{Results on Trans101}
On Trans101, SEKI demonstrates outstanding performance across multiple tasks, particularly in the Cls.O., Auto, Normal, Room, and Jigsaw tasks, with an average rank of 7.7 across all tasks, achieving the best ranking among all methods and proving its efficacy in multi-task optimization. 

In summary, SEKI demonstrates outstanding performance across datasets, with strong optimization efficiency, accuracy, and multi-task handling capabilities, making it a promising approach for future neural architecture search. 

\subsection{Ablation Study}
\label{exp:as}
In this section, we ablate different hyper-parameters of SEKI. Specifically, we first search for the optimal allocation of iterations between the self-evolution and knowledge inspiration stages. Then, we explore the best settings for $k$ and $\xi$, i.e, the number of the best-selected candidates and the number of the randomly selected candidates among $k$, respectively, in the knowledge inspiration phase. We evaluate the results on CIFAR-10.
Finally, we study the performance of SEKI with different LLMs. All experiments are conducted on DARTS search space.


\subsubsection{Self-Evolution vs Knowledge Inspiration}
\Cref{table:ab1} compares the performance of SEKI under different self-evolution coefficients $\lambda$ and knowledge inspiration coefficients $\gamma$, with a fixed total number of iterations, i.e., $\lambda + \gamma = 50$, following the literature. As shown in the table, when $\lambda$ is small, the number of candidate networks in the knowledge repository is limited, and each candidate undergoes insufficient self-evolution, resulting in low performance. Consequently, the performance is inferior. As $\lambda$ increases, performance gradually improves. However, when $\lambda$ becomes large, although the number of candidates in the knowledge repository increases and some candidates experience sufficient self-evolution, the round of knowledge inspiration iterations becomes constrained, leading to inferior performance once again. The best performance is achieved at $\lambda = 35$, indicating that both self-evolution and knowledge inspiration are crucial and effectively contribute to the performance of SEKI.

\subsubsection{$k$ and $\xi$ in Knowledge Inspiration}
\Cref{table:ab21} demonstrates the impact of $k$ and $\xi$ in knowledge inspiration stage on SEKI performance, where $k$ represents the number of top architectures selected in the repository, and $\xi$ is the number of architectures randomly chosen from these $k$ to construct the input. Adjusting these two variables optimizes the ability of LLM to inspire better architectures. 

From the table, when both $k$ and $\xi$ are too small, the diversity of the knowledge repository is limited, impairing the LLM’s ability to extract effective design patterns, which in turn hampers search performance. On the other hand, when both parameters are too large, they may introduce more suboptimal architectures in search, reducing the overall input quality and negatively impacting the LLM's output. The best performance is achieved at $k=16$ and $\xi=k/2$ (i.e., $\xi=8$), which strikes the optimal balance between the diversity of the knowledge repository and the quality of input prompts, effectively enhancing the LLM’s ability to generate high-quality architectures.

\subsubsection{Different LLMs}
We compare the performance of SEKI using Qwen2.5-32B \cite{qwen2.5} and GPT4o-mini \cite{openai2024gpt4ocard}, across CIFAR-10, CIFAR-100, and ImageNet-1K in \Cref{table:basemodel}. On CIFAR-10 and CIFAR-100, SEKI-Qwen slightly outperforms SEKI-GPT, while on ImageNet-1K SEKI-GPT achieves slightly better performance. Overall the differences are subtle, indicating that SEKI is robust and generalizable to different LLMs. 
\begin{table}[ht]
\centering
\footnotesize
\caption{Comparison of using GPT and Qwen as the LLM on SEKI. Top-1 accuracy is measured.}

\begin{tabular}{l|ccc}
\hline
\multirow{2}{*}{Base model} & CIFAR-10 & CIFAR-100 & ImageNet-1K \\ \cline{2-4} 
 & \begin{tabular}[c]{@{}c@{}}Test Acc\\ (\%)\end{tabular} & \begin{tabular}[c]{@{}c@{}}Test Acc\\ (\%)\end{tabular} & \begin{tabular}[c]{@{}c@{}}Test Acc\\ (\%)\end{tabular} \\ \hline
SEKI-Qwen & 97.71 & 84.14 & 75.58 \\
SEKI-GPT & 97.67 & 83.93 & 75.68 \\ \hline
\end{tabular}
\label{table:basemodel}
\end{table}
\vspace{-4mm}

\section{Conclusion}

In this paper, we propose SEKI, a novel LLM-based NAS method that leverages the self-evolution and knowledge distillation capabilities of LLMs to iteratively optimize network search strategies—without requiring any domain-specific data. SEKI achieves state-of-the-art (SOTA) performance across various datasets, search spaces, and tasks, while requiring only 0.05 GPU-days for the search process. This demonstrates the effectiveness, efficiency, and generalization of our approach. While SEKI is simple, it demonstrates the success of LLM in NAS research. Exploring advanced algorithms with LLM could be the next step. Additionally, extending SEKI to a broader range of domains and tasks presents an exciting direction for future research.

\section{Impact Statement}
This paper presents work whose goal is to advance the field of Machine Learning. There are many potential societal consequences of our work, none which we feel must be specifically highlighted here.


\bibliographystyle{plainnat}
\bibliography{nips2024.bib}

\newpage
\appendix
\onecolumn
\section{Appendix}

While this paper does not primarily focus on it, we conduct a small test on the Penn Treebank (PTB) corpus \cite{10.5555/972470.972475}, a benchmark dataset for natural language processing (NLP) tasks, to evaluate the performance of our SEKI method in a different domain. \Cref{table:ptb} compares SEKI with other leading neural architecture search (NAS) methods. SEKI achieves the lowest test perplexity of 55.69, outperforming other methods while using only 22 M parameters, demonstrating both high computational efficiency and strong performance. Moreover, this result highlights SEKI’s strong generalization capabilities across different tasks.

\begin{figure}[h]
  \centering 
    \includegraphics[width=0.9\textwidth]{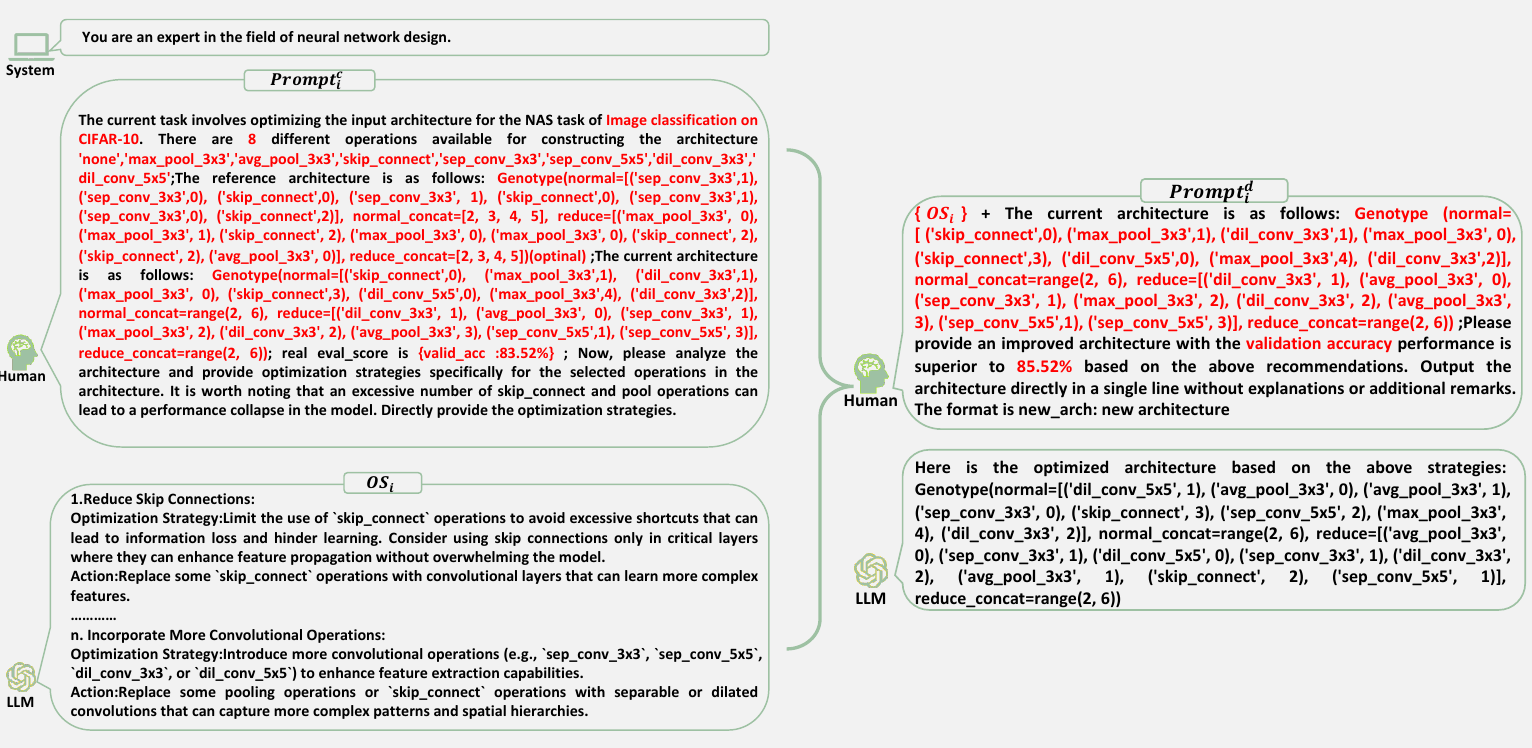}
  \caption{An example of Self-Evolution.}
  \label{fig:pp1}
\end{figure}

\begin{figure}[t]
  \centering 
    \includegraphics[width=0.6\textwidth]{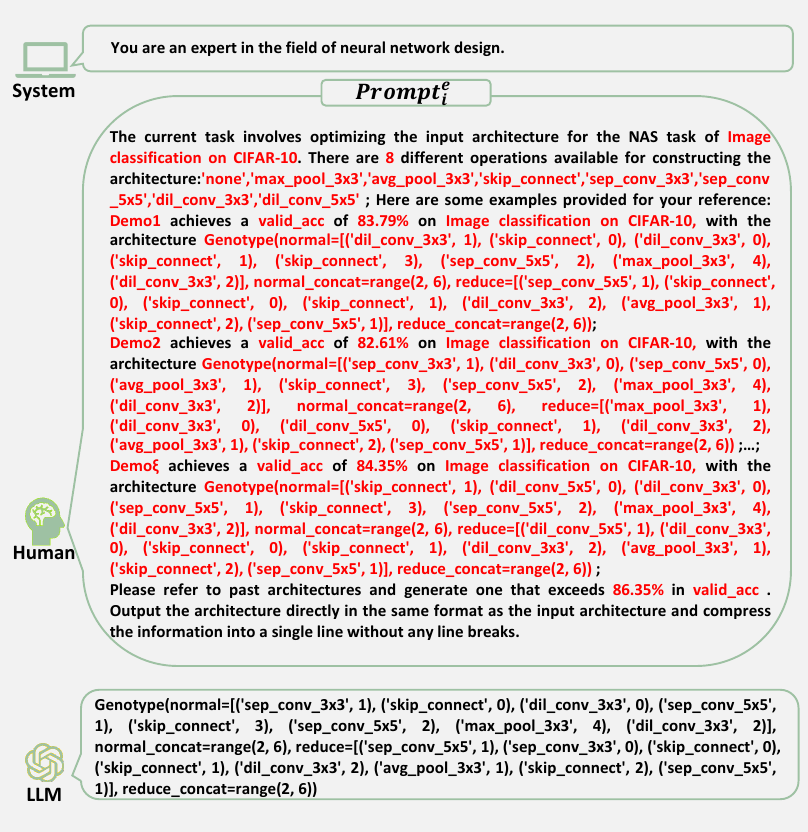}
  \caption{An example of Knowledge Inspiration.}
  \label{fig:pp2}
\end{figure}

\begin{figure}[ht]
  \centering 
    \includegraphics[width=0.9\textwidth]{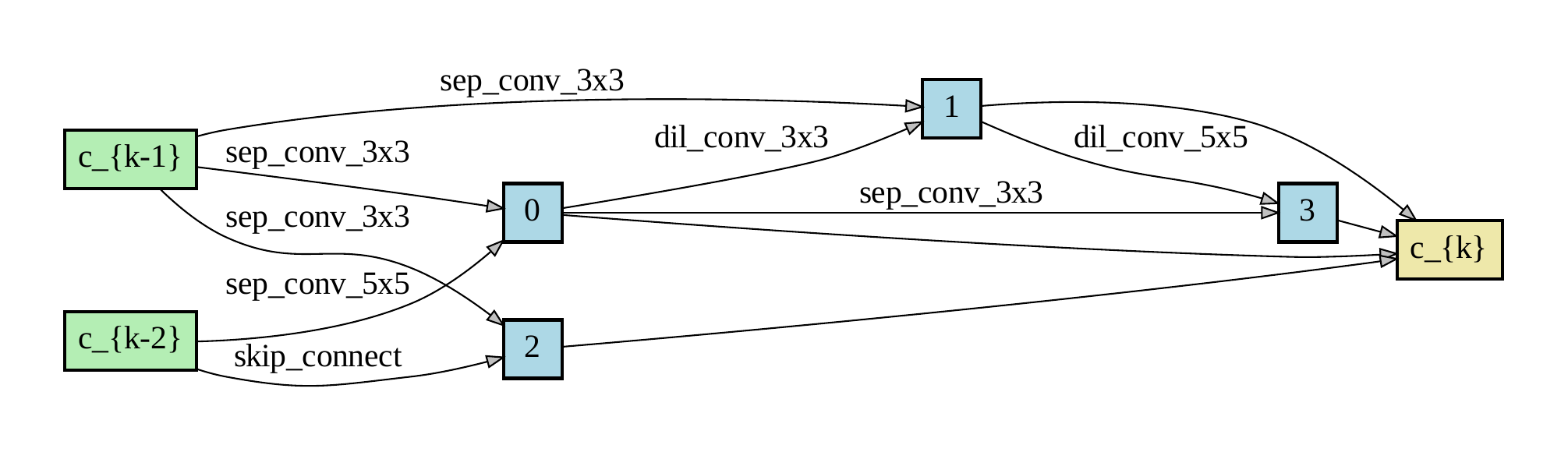}
  \caption{SEKI normal cell learned on CIFAR-10.}
  \label{fig:nor1}
\end{figure}

\begin{figure}[ht]
  \centering 
    \includegraphics[width=0.9\textwidth]{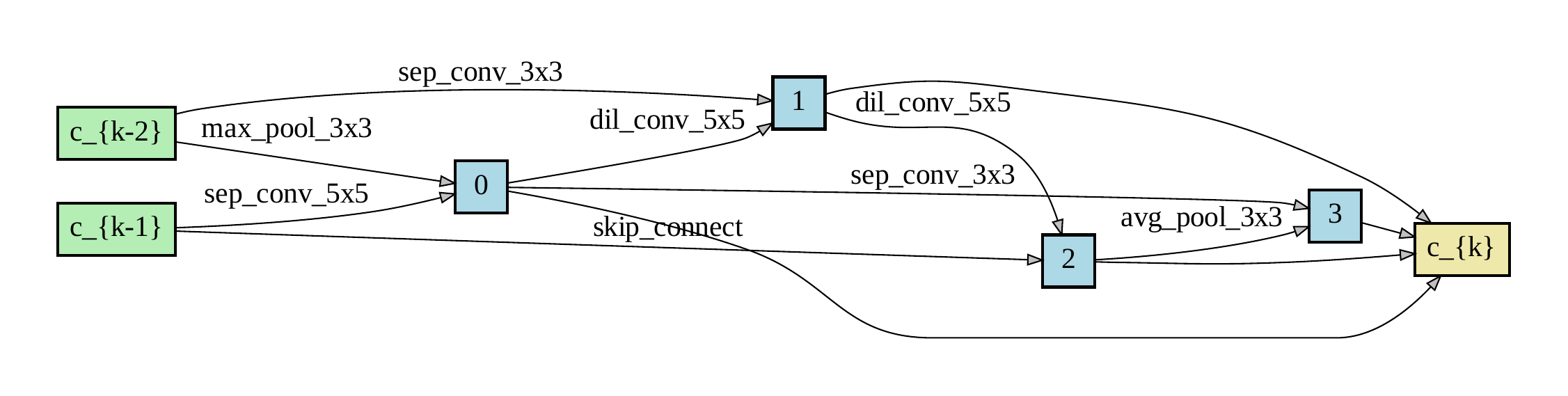}
  \caption{SEKI reduction cell learned on CIFAR-10.}
  \label{fig:red1}
\end{figure}

\begin{figure}[ht]
  \centering 
    \includegraphics[width=0.9\textwidth]{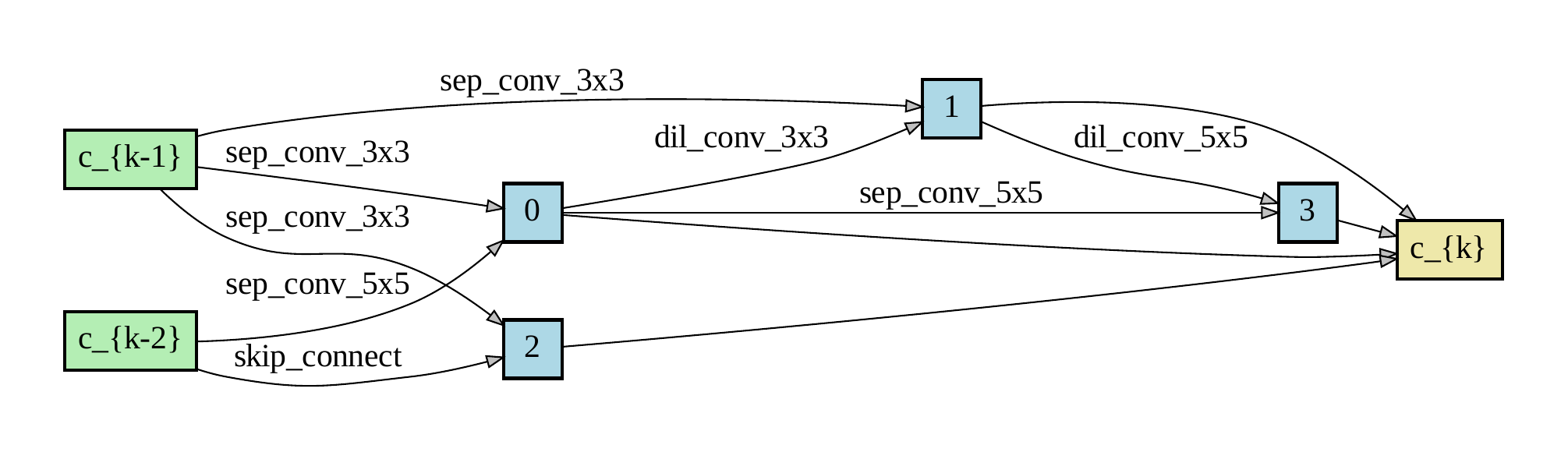}
  \caption{SEKI normal cell learned on ImageNet-1K.}
  \label{fig:nor2}
\end{figure}

\begin{figure}[ht]
  \centering 
    \includegraphics[width=0.9\textwidth]{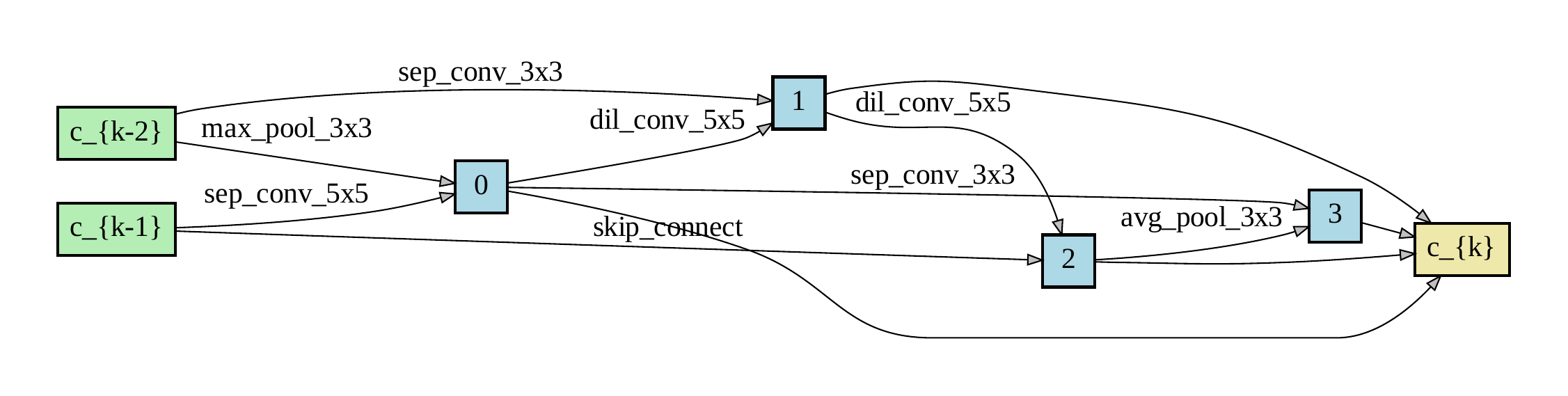}
  \caption{SEKI reduction cell learned on ImageNet-1K.}
  \label{fig:red2}
\end{figure}

\Cref{fig:pp1} and ~\Cref{fig:pp2} present examples of self-evolution and knowledge inspiration. ~\Cref{fig:nor1} to ~\Cref{fig:red2} present searched architectures on CIFAR-10 and ImageNet-1K, respectively, on DARTS search space.


\begin{table}[h]
\centering
\footnotesize
\caption{Performance comparison on PTB.}
\begin{tabular}{lcccc}
\hline
\multicolumn{1}{c}{\multirow{2}{*}{Architecture}} & \multicolumn{2}{c}{Perplexity(\%)} & \multirow{2}{*}{\begin{tabular}[c]{@{}c@{}}Params\\ (M)\end{tabular}} & \multirow{2}{*}{Search Method} \\ \cline{2-3}
\multicolumn{1}{c}{} & valid & test &  &  \\ \hline
LSTM + SE & 58.1 & 56.0 & 22 & manual \\ \hline
NAS & - & 64.0 & 25 & RL \\
ENAS & 60.8 & 58.6 & 24 & RL \\
DARTS(1st) & 60.2 & 57.6 & 23 & GD \\
DARTS(2st) & 58.1 & 55.7 & 23 & GD \\
GDAS & 59.8 & 57.5 & 23 & GD \\
NASP & 59.9 & 57.3 & 23 & GD \\
SDARTS-RS & 58.7 & 56.4 & 23 & GD \\
SDARTS-ADV & 58.3 & 56.1 & 23 & GD \\ \hline
SEKI & 58.1 & 55.69 & 22 & LLM \\ \hline
\end{tabular}
\label{table:ptb}
\end{table}



\end{document}